\documentclass[11pt]{article}

\usepackage{graphicx} % For images
\usepackage{subfigure}

\usepackage{amsmath, amssymb} % For math symbols
\usepackage{algorithm}
\usepackage{float}
\usepackage{algorithmic}
\usepackage[numbers]{natbib} % Citation style compatible with arXiv
\usepackage{authblk} % For author and affiliation formatting
\usepackage[margin=1in]{geometry} % Adjust margins for arXiv
\usepackage{hyperref}
\usepackage{url}
\usepackage{booktabs}
\usepackage{multirow}
\usepackage{bm}
\usepackage{booktabs} % for professional tables

\newtheorem{theorem}{Theorem}[section]

\newtheorem{lemma}[theorem]{Lemma}

\title{EFiGP: Eigen-Fourier Physics-Informed Gaussian Process for Inference of Dynamic Systems}
\author[1]{Jianhong Chen}
\author[2]{Shihao Yang\thanks{Corresponding Author: \texttt{shihao.yang@isye.gatech.edu}}}

\affil[1]{Department of Mechanical \& Industrial Engineering, Northeastern University, Boston, MA, USA}
\affil[2]{H. Milton Stewart School of Industrial and Systems Engineering, Georgia Institute of Technology, Atlanta, GA, USA}

\begin{document}
\maketitle
\begin{abstract}
    Parameter estimation and trajectory reconstruction for data-driven dynamical systems governed by ordinary differential equations (ODEs) are essential tasks in fields such as biology, engineering, and physics. These inverse problems -- estimating ODE parameters from observational data -- are particularly challenging when the data are noisy, sparse, and the dynamics are nonlinear. We propose the Eigen-Fourier Physics-Informed Gaussian Process (EFiGP), an algorithm that integrates Fourier transformation and eigen-decomposition into a physics-informed Gaussian Process framework. This approach eliminates the need for numerical integration, significantly enhancing computational efficiency and accuracy. Built on a principled Bayesian framework, EFiGP incorporates the ODE system through probabilistic conditioning, enforcing governing equations in the Fourier domain while truncating high-frequency terms to achieve denoising and computational savings. The use of eigen-decomposition further simplifies Gaussian Process covariance operations, enabling efficient recovery of trajectories and parameters even in dense-grid settings. We validate the practical effectiveness of EFiGP on three benchmark examples, demonstrating its potential for reliable and interpretable modeling of complex dynamical systems while addressing key challenges in trajectory recovery and computational cost.
\end{abstract}

\section{Introduction}

Systems of coupled Ordinary Differential Equations (ODEs) are essential tools for modeling the intricate mechanisms underlying various scientific and engineering processes, such as neuroscience \citep{Fitzhugh1961, Nagumo1962}, ecology \citep{Lotka1932}, and systems biology \citep{Hirata2002}.

We focus on dynamical systems governed by the following ODE formulation, as studied in \citet{yang2021a, seifner2024foundational, gorbach2017scalable}:
\begin{equation}
    \dot{\bm{x}}(t) = \frac{d\bm{x}(t)}{dt} = f(\bm{x}(t), \bm{\theta}, t), \quad t \in [0, T],
\label{ode}
\end{equation}
where the vector \( \bm{x}(t) \) represents the system outputs that change over time \( t \), and \( \bm{\theta} \) is the vector of model parameters to be estimated from experimental or observational data. When the function \( f \) is nonlinear, determining \( \bm{x}(t) \) given the initial conditions \( \bm{x}(0) \) and \( \bm{\theta} \) typically requires a numerical integration method, such as Runge–Kutta \citep{lapidus1971}.

Traditionally, ODEs have been utilized more for conceptual or theoretical insights rather than for fitting data, due to limitations in the availability of experimental data. However, advancements in experimental and data collection techniques have enhanced the ability to monitor dynamical systems in near real-time. Typically, such data are recorded at discrete time points and are subject to measurement errors. Therefore, we assume that the observations \( \bm{y}(\bm{\tau}) = \bm{x}(\bm{\tau}) + \bm{\epsilon}(\bm{\tau}) \) are made at \( N \) specific time points \( \bm{\tau} = (\tau_1, \tau_2, \ldots, \tau_N) \), with the error \( \bm{\epsilon}(\bm{\tau}) \) governed by a noise level \( \sigma \). Our focus is on inferring the parameters \( \bm{\theta} \) and recovering the ground-truth trajectory $\{\bm{x}(t)\}_{t=0}^T$ given data \( \bm{y}(\bm{\tau}) \), with particular emphasis on the nonlinear structure of \( f \).

\textbf{Background}\quad Bayesian inference and Gaussian Processes have long been utilized for calibrating parameters in dynamical systems \citep{kennedy2001bayesian}. More recently, MAnifold-constrained Gaussian process Inference (MAGI) for ODEs \citep{yang2021a} and Physics-Informed Gaussian Process (PIGP) for Partial Differential Equations (PDEs) \citep{li2024parameter} have emerged as principled Bayesian approaches that inherently incorporate physical information to estimate parameters from observational data. These Bayesian counterparts to the Physics-Informed Neural Network (PINN) \citep{raissi2019} provide native uncertainty quantification within a theoretically rigorous Bayesian framework. MAGI and PIGP achieve this by leveraging a physics-informed Bayesian conditioning mechanism, which constrains the difference between derivative information obtained from the governing differential equations and that derived from a Gaussian Process (GP).

Focusing on ODEs, one of the key advantages of MAGI is that it bypasses numerical integration, leading to high computational efficiency with strong empirical performance. However, while the Bayesian conditioning of a physics-informed, ODE-driven manifold constraint provides a theoretically ideal inference method, its practical implementation requires discretization. This discretization introduces some degree of inaccuracy due to approximation errors. The discretized physics-informed constraint can be viewed as a collocation method, where the ODE information is conditioned only on specific collocation points. As a result, the computational burden of MAGI increases linearly with the number of discretization points. In this paper, we address this challenge by transforming the physics-informed, ODE-driven manifold constraint into Fourier space and applying spectral decomposition to the GP quadratic form. These techniques are employed to further reduce the computational cost associated with MAGI.

\textbf{Literature Review}\quad The problem of inferring dynamical systems described by ODEs has been extensively studied. Traditional approaches, such as those proposed by \citet{bard1975} and \citet{benson1979}, typically involve estimating parameters by minimizing the deviation between observed data and system responses through numerical integration. However, these methods are computationally expensive due to the repeated evaluations required. To mitigate these challenges, smoothing techniques have been developed, such as the penalized likelihood approach by \citet{ramsay2007}, which employs B-splines to construct functions that fit observed data while satisfying the ODE system constraints. In contrast to penalized likelihood methods, Gaussian Processes provide a flexible and analytically tractable alternative for smoothing within a Bayesian framework, as demonstrated by \citet{hennig2015}.

Previous studies \citep{calderhead2009, dondelinger2013, wang2014, wenk2019} have explored the use of GP to approximate dynamical systems, but these methods faced issues of conceptual incompatibility. The introduction of MAnifold-constrained Gaussian process Inference (MAGI) by \citet{yang2021a} and \citet{wenk2020} addressed these challenges by resolving the theoretical incompatibilities of earlier GP-based approaches. MAGI not only improves the accuracy of parameter inference but also achieves computational efficiency, with its runtime scaling linearly with the number of system components \citep{wenk2020}.

Physics-Informed Neural Network (PINN) \citep{raissi2019} offers another realm for solving differential equations using machine learning. By embedding the governing physical laws directly into the network’s loss function, PINN can effectively handle high-dimensional and nonlinear PDEs without requiring large datasets, as the physics loss guides the optimization. However, despite their versatility, PINN can be computationally expensive and prone to failure, especially in multi-scale dynamical systems, due to challenges such as stiff gradients and sensitivity to hyperparameters \citep{wang2021understanding}. 

To address these limitations, \citet{li2021} introduced the Fourier Neural Operator (FNO), which leverages Fourier transforms to solve differential equations by operating in the Fourier domain. In this space, differentiation simplifies multiplication, enabling FNO to efficiently capture long-range dependencies and complex interactions with quasi-linear time complexity. While FNO has demonstrated state-of-the-art approximation capabilities, they still require tens of thousands of training pairs generated by numerical solvers.

Given the computational limitations and performance instability of these approaches, there is a clear need for methods that are both robust and efficient. Notably, no prior work has explored incorporating Fourier transforms into Bayesian Gaussian Process frameworks that completely bypass numerical solvers, presenting an opportunity for innovation in this domain. 

\textbf{Our Contribution}\quad We propose a novel algorithm incorporating truncation with Fourier Transformation and Eigen-decomposition in the Physics-informed Gaussian Process (EFiGP). We demonstrate that the resulting parameter inference and trajectory recovery are statistically sound, computationally efficient, and effective in various practical scenarios. Our EFiGP not only addresses the limitations of previous Gaussian process approaches when discretization becomes very dense, but also improves accuracy and reduces computation time.

In particular, the physics information from the governing equation is enforced in the Fourier domain, which is especially useful for oscillatory ODEs that describe periodic or quasi-periodic systems. Examples include biological rhythms, such as the oscillations of Hes1 mRNA and Hes1 protein \citep{Hirata2002}, and relaxation oscillators, such as the FitzHugh-Nagumo equations \citep{Fitzhugh1961}. Additionally, our approach allows for the truncation of high-frequency terms in the Fourier-transformed ODEs representation, effectively achieving denoising while reducing computational costs.

The incorporation of Eigen-decomposition truncation in our algorithm enhances the computational efficiency and accuracy of parameter estimation and trajectory recovery. Eigen-decomposition allows us to diagonalize the covariance matrix of the Gaussian Process, which simplifies the computational complexity involved in the multiplication of large matrices. This is particularly advantageous in high-discretization settings, where points in recovered trajectories are highly correlated due to smoothness. By breaking down the Gaussian Process into orthogonal components and truncation, our approach efficiently captures the essential features of the dynamical system trajectories while discarding redundant information. This decomposition not only accelerates the computation but also enhances numerical stability, reducing the risk of errors due to ill-conditioned matrices. Lastly, a Python implementation of EFiGP is provided on GitHub\footnote{\url{https://github.com/PeChen123/EFiGP}} for public use
\section{Preliminaries}

\subsection{ODE Inverse Problem}

Typically, an ODE model is expressed as Eq.\eqref{ode}. In many practical applications, we are faced with the challenge of determining the underlying parameters \( \bm{\theta}\) from observed data. This leads to the ODE inverse problem, which can be formally stated as follows.

Given a set of observed data points \(\{(\tau_i, y(\tau_i))\}_{i=1}^N\), determine the parameter \( \bm{\theta} \) such that the solution \( \bm{x}(\bm{\tau}; \bm{\theta}) \) of the ODE in  Eq.\eqref{ode} best fits the observed data in the sense of minimizing the discrepancy between the ODE solution and the observations.

The inverse problem is often approached through optimization techniques, where an objective function, usually the sum of squared differences between the observed and the ODE solution, is minimized:

\[
\min_{\bm{\theta}} \sum_{i=1}^N \| x(\tau_i; \bm{\theta}) - y(\tau_i) \|^2.
\]

Each evaluation of loss function typically requires solving $x(\tau_i; \bm{\theta})$ using numerical integration.

\subsection{MAGI: Manifold-Constrained Gaussian Process Inference}

The MAGI framework, introduced by \citet{yang2021a}, establishes a Bayesian approach to solve inverse problems using the Gaussian process. For a concise overview of the Gaussian process, see supplementary materials \S\ref{GPSm}. Within this Bayesian framework, the \(D\)-dimensional dynamical system \(\bm{x}(t)\) is modeled as a realization of the stochastic process \(\bm{X}(t) = (X_1(t), X_2(t), \ldots, X_D(t))\), with the model parameters \(\bm{\theta}\) represented as realizations of the random variable \(\bm\Theta\). The posterior distribution is then naturally derived. For clarity and conciseness, the main text omits the subscript \(d\) corresponding to each dimension of the ODE system, with the complete \(d\) notation detailed in the supplementary material~\S\ref{fulld}.

\textbf{Prior:} A general prior \(\pi(\cdot)\) is imposed on \(\bm{\theta}\), and an independent GP prior is assumed for each component \(\bm{X}(t)\), such that
\begin{equation}
    \bm{X}(t) \sim \mathcal{GP}(\mu, \mathcal{K}) \quad t \in [0, T],
\end{equation}
where the mean function \(\mu : \mathbb{R} \to \mathbb{R}\) and the positive-definite covariance function \(\mathcal{K} : \mathbb{R} \times \mathbb{R} \to \mathbb{R}\) are parameterized by hyperparameters \(\bm{\phi}\). Therefore, for any finite set of time points \(\bm{\tau}\), \(\bm{X}(\bm{\tau})\) follows a multivariate Gaussian distribution with mean vector \(\bm{\mu}(\bm{\tau})\) and covariance matrix \(\mathcal{K}(\bm{\tau}, \bm{\tau})\).

\textbf{Likelihood:}  Let the observations be denoted by \(\bm{y}(\bm{\tau}) = (y(\tau_1), \ldots, y(\tau_N))\), where \(\bm{\tau} = (\tau_1, \ldots, \tau_N)\) represents the set of \(N\) observation time points for each component. For simplicity, the observation noise for each component is assumed to be i.i.d. zero-mean Gaussian with variance \(\sigma^2\). Thus, the observation likelihood is given by:
\begin{equation}
    \bm{Y}(\bm{\tau}) \mid \bm{X}(\bm{\tau}) = \bm{x}(\bm{\tau}) \sim \mathcal{N}(\bm{x}(\bm{\tau}), \sigma^2 \bm{I}_{N})
\end{equation}

\textbf{Physics Information:} A new random variable is introduced to quantify the difference between the time derivative \(\dot{\bm{X}}(t)\) of the GP and the ODE structure for a given value of the parameter \(\bm{\theta}\):
\begin{equation}
    W = \sup_{t \in [0, T]} \left|{\dot{\bm{X}}(t)} - f(\bm{X}(t), \bm{\theta}, t) \right|
\end{equation}
Under the event \(\{W = 0\}\), the stochastic process \(\bm{X}(t)\) fully satisfies ODE function Eq.\eqref{ode}. Therefore, conditioning on $W=0$ will impose a physics-informed Bayesian constraint on the GP, \(\bm{X}(t)\). However, since \(W\) is a supremum over an uncountable set, it cannot be computed analytically. To address this, an approximation \(W_{I}\) based on a finite discretization of the set \(I = (t_1, t_2, \ldots, t_n)\) with \(n\) discretization points is used, such that \(\bm{\tau} \subset I \subset [0, T]\):
\begin{equation}
    W_{I} = \max_{t \in I} \left|{\dot{\bm{X}}(t)} - f(\bm{X}(t), \bm{\theta}, t) \right|
\end{equation}

Note that \(W_{I}\) is the maximum of a finite set, and \(W_{I}\) converges monotonically to \(W\) as \(I\) becomes denser within \([0, T]\). Consequently, this allows for the analytical derivation of a posterior distribution.

\textbf{Posterior:} The practically computable posterior distribution is given by
\begin{equation}
    p_{\bm{\Theta}, \bm{X}(I) \mid W_I, \bm{Y}(\bm{\tau})}(\bm{\theta}, \bm{x}(I) \mid W_I = 0, \bm{Y}(\bm{\tau}) = \bm{y}(\bm{\tau}))
\end{equation}
which represents the joint conditional distribution of \(\bm{\theta}\) and \(\bm{X}(I)\). This formulation allows for the simultaneous inference of both the parameters \(\bm{\theta}\) and the unobserved trajectory \(\bm{X}(I)\) from the noisy observations \(\bm{y}(\bm{\tau})\). Using Bayes’ rule, this posterior can be expressed as:
\begin{equation}
    \begin{aligned}
    p_{\bm\Theta, \bm{X}(I) \mid W_I, \bm{Y}(\bm{\tau})}(\bm{\theta}, \bm{x}(I) \mid W_I = 0, \bm{Y}(\bm{\tau}) = \bm{y}(\bm{\tau})) &\propto \\
    P(\bm\Theta = \bm{\theta}, \bm{X}(I) = \bm{x}(I), W_I = 0, \bm{Y}(\bm{\tau}) = \bm{y}(\bm{\tau})).
\end{aligned}
\end{equation}
The right-hand side can be expressed in closed form as follows:
\begin{equation}
    \begin{aligned}
    &P(\bm\Theta = \bm\theta, \bm{X}(I) = \bm{x}(I), W_I = 0, \bm{Y}(\bm{\tau}) = \bm{y}(\bm{\tau})) \\
    &= \pi_{\bm\Theta}(\bm\theta) \times P(\bm{X}(I) = \bm{x}(I) \mid \bm\Theta = \bm\theta ) \\
    &\times P(\bm{Y}(\bm{\tau}) = \bm{y}(\bm{\tau}) \mid \bm{X}(I) = \bm{x}(I), \bm\Theta = \bm\theta) \\
    &\times P(W_I = 0 \mid \bm{Y}(\bm{\tau}) = \bm{y}(\bm{\tau}), \bm{X}(I) = \bm{x}(I), \bm\Theta = \bm\theta).
\end{aligned}
\label{MAGI}
\end{equation}

We will briefly review spectral decomposition and Fourier transformation before the further EFiGP derivation and approximation of this posterior function in the next section.

\subsection{Spectral Decomposition}

To sample from a low-dimensional space for multivariate Gaussian distributions, the spectral decomposition (eigendecomposition) method can be applied. 

\begin{lemma}
    Let \(\Sigma\) be a covariance matrix with eigendecomposition \(\Sigma = V \Lambda V^{\top}\), where \(V\) is the matrix of eigenvectors, \(\Lambda\) is the diagonal matrix of eigenvalues, and \(J\) is the number of non-zero eigenvalues. Consider a random vector \(\bm{Z} \sim \mathcal{N}(0, I_J)\), where \(I_J\) is the \(J \times J\) identity matrix. Then, the distribution of \(\mu + V\Lambda^{1/2} \bm{Z}\) is \(\mathcal{N}(\mu, \Sigma)\).

    This can be equivalently expressed as a sum:
    \begin{equation}
            \mu + \sum_{i=1}^{J}\sqrt{\lambda_i} z_i \bm{v}_i \sim \mathcal{N}(\mu, \Sigma),
    \end{equation}

    where \(\lambda_i\) are the eigenvalues in \(\Lambda\), \(\bm{v}_i\) are the corresponding eigenvectors in \(V\), and \(z_i\) are the components of the random vector \(\bm{Z}\). 
\label{Eigen}
\end{lemma}

This formulation also allows for truncation, where small eigenvalues are ignored to reduce computational complexity while maintaining a close approximation to the original distribution.

\subsection{Fourier Transformation}

Recall that the Discrete Fourier Transform (DFT) is a type of linear transformation that can be represented by a matrix, denoted as \(A_{\text{DFT}}\). This matrix acts on vectors in \(\mathbb{R}^n\), transforming them into vectors in \(\mathbb{C}^n\).

To work within a real-valued framework, we consider a linear mapping from \(\mathbb{R}^{n}\) to \(\mathbb{R}^{2n}\) that augments the DFT output by separating the real and imaginary components, represented by \(\tilde{A}\). Specifically, this Fourier operator is constructed by combining the DFT matrix with a process that augments the output into its real and imaginary parts.

For a detailed analysis of this mapping, including the mathematical derivation and implications, please refer to the supplemental material (see \S~\ref{transma}).

By the properties of the Gaussian distribution, we have the following result:

\begin{lemma}
The DFT and augmentation of Gaussian random vector \(\bm{X}(I) \sim \mathcal{N}(\bm{\mu}_I, K_{I,I})\) also results in a multivariate Gaussian distribution. If we use $\mathcal{F}$ to denote Fourier transform and subsequent separation of real part and imaginary part, then \(\mathcal{F}\{\bm{X}(I)\} = \tilde{A}\bm{X}(I) \sim \mathcal{N}(\tilde{A}\bm{\mu}_I, \tilde{A} \cdot K_{I,I} \cdot \tilde{A}^\top)\), where \(\tilde{A}\) is the combined matrix form of the discrete Fourier transform and the augmentation process that separates the real and imaginary parts.
\label{DFT}
\end{lemma}

Note that the resulting covariance matrix after applying this transformation is of dimension \(\mathbb{R}^{2n-1 \times 2n-1}\), and the mean vector is in \(\mathbb{R}^{2n-1}\). The detailed closed form of mapping matrix $\tilde{A}$ is included in the supplemental material \S\ref{transma}. This formulation also allows for truncation, where high-frequency terms are ignored to again reduce computational complexity while maintaining a close approximation to the original distribution.

\section{EFiGP: Eigen-Fourier Physics-Informed Gaussian Process}

We tackle two limitations of the previous Gaussian process approach: When the discretization set becomes very dense, (1) the computational cost increases significantly, and (2) the algorithm may fail to converge to the ODEs solution trajectory due to highly correlated posterior samples. Thus, we combine the ideas of eigen-decomposition and Fourier transformation to reduce computational cost in sampling and improve the accuracy of the random variable \(W\) characterization of ODEs and GP discrepancy. 

\textbf{Fourier:} We now measure the deviation of GP and the ODEs requirement in the Fourier space:

\begin{equation}
W_I^\mathcal{F} = \max \left|\mathcal{F}[\dot{\bm{X}}(I)] - \mathcal{F}[f(\bm{X}(t),\bm{\theta},I)] \right|
\end{equation} 
where the set \(I = (t_1, t_2, \ldots, t_n)\) with \(n\) discretization points. We can truncate the discrete Fourier series at the  \(l\)-th term (\( l < n\)) to reduce the computational cost. Furthermore, we can easily obtain the computational form by Lemma~\ref{DFT} since \(\dot{\bm{X}}\) is a joint Gaussian distribution (see \citet{rasmussen2006}, chapter 9) as 

\begin{align*}
& P(W_I^\mathcal{F} = 0 \mid \bm{Y}(\bm{\tau}) = \bm{y}(\bm{\tau}), \bm{X}(I) = \bm{x}(I), \bm\Theta = \bm\theta) \\
=& P( \mathcal{F}[\dot{\bm{X}}(I)] = \mathcal{F}[f(\bm{x}(t),\bm{\theta},I)]  \mid \bm{X}(I) = \bm{x}(I)) \\
\propto &\exp \{ -\frac{1}{2}\left\|\tilde A_{(l)}\{ f^{\bm{\theta},\bm{x}}_{I} - '\mathcal{K}(I,I)\mathcal{K}(I,I)^{-1}\{ \bm{x}(I) - \bm{\mu}(I) \}\} \right\|^2_{(C_{(l)}^{\mathcal{F}})^{-1}} \}
\end{align*}

where \(f^{\bm{\theta},\bm{x}}_{I}\) is short notation for \(f(\bm{x}(I), \bm{\theta}, I) \), and $\|\cdot\|^2_\cdot$ is short notation for quadratic form $\|\bm{r}\|^2_B = \bm{r}^T B \bm{r}$. The matrix $\tilde A_{(l)}$ is the truncated Fourier transform matrix at $l$-th frequency term, and \(C_{(l
)}^{\mathcal{F}} = \tilde A_{(l)}\cdot C\cdot \tilde A_{(l)}^\top\) can be obtained by the Lemma~\ref{DFT} on the conditional covariance matrix $C = \mathcal{K}''(I,I)-'\mathcal{K}(I,I)\mathcal{K}(I,I)^{-1}\mathcal{K}'(I,I)$. Finally, \( '\mathcal{K} = \frac{\partial}{\partial{s}}\mathcal{K}(s,t)\), \(\mathcal{K}' = \frac{\partial}{\partial{t}}\mathcal{K}(s,t)\), and \(\mathcal{K}'' = \frac{\partial^2}{\partial{s}\partial{t}}\mathcal{K}(s,t)\). All the closed forms can be found in supplementary material~\S\ref{transma}.

\textbf{Eigen:} Since posterior sampling or maximum a posteriori (MAP) optimization on \(\bm{X}(I)\) in Eq.\eqref{MAGI} incurs a high cost when the set becomes denser, we propose an efficient way to handle \(\bm{X}(I)\) by using spectral decomposition (Lemma~\ref{Eigen}). We consider the change of variable (orthogonally reparametrize) \(\bm{X}(I)\) to \(\bm{z} = (z_1, \ldots, z_n)\), using the matrix square root from the spectral decomposition of the prior variance and covariance matrix:
\begin{equation}
\bm{X}(I) = \bm{\mu}(I) + V_{(j)}\Lambda_{(j)}^{\frac{1}{2}}\bm{z} = \bm{\mu}(I) + \sum_{i=1}^{j} z_i \sqrt{\lambda_i} \bm{v}_i
\label{COV}
\end{equation}

where \(\lambda_i, \bm{v}_i\) are eigenvalues and eigenvectors of \(\mathcal{K}(I, I)\). The matrices \(V_{(j)} = (\bm{v}_1, \ldots, \bm{v}_j)\), \(\Lambda_{(j)} = \mathrm{diag}(\lambda_1, \ldots, \lambda_j)\) are truncated eigen decomposition at $j$-th eigen value term. The $j$ is a truncation number hyper-parameter that aims to save computational cost over all parts of the objective function Eq.\eqref{MAGI}.

\textbf{Posterior:} Now, our new practically computable posterior distribution for Eigen-Fourier Physics-Informed Gaussian Process (EFiGP) is:

\begin{equation}
\label{obj}
\begin{split}
& P(\bm\Theta = \bm{\theta}, \bm{Z} = \bm{z} \mid W^\mathcal{F}_I = 0, \bm{Y}(\tau)= \bm{y}(\tau))  \\
\propto  & P(\bm\Theta = \bm{\theta}, \bm{X}(I) = \bm{x}(I), W^\mathcal{F}_I = 0, \bm{Y}(\tau)= \bm{y}(\tau))\times J(\bm{X}(I) \to \bm{Z})\\
\propto &\ \pi_{\bm\Theta}(\bm{\theta}) \exp \left\{ -\frac{1}{2} \left[  \bm{z}^T\bm{z} \right.\right. \\
&+\left\|\tilde A_{(l)}\cdot f^{\bm{\theta},\bm{x}}_{I} - \tilde A_{(l)} \cdot '\mathcal{K}(I,I)\mathcal{K}(I,I)^{-1}\{V_{(j)}\Lambda_{(j)}^{\frac{1}{2}}\bm{z}\} \right\|^2_{(C_{(l)}^{\mathcal{F}})^{-1}}\\
&+ \left.\| \bm{\mu}(I) + V_{(j)}\Lambda_{(j)}^{\frac{1}{2}}\bm{z} - \bm{y} \|^2_{\sigma^{-2}} \left. \right] \right\}
\end{split}
\end{equation}

where $\bm{x}({I}) = \bm{\mu}(I) + V_{(j)}\Lambda_{(j)}^{\frac{1}{2}}\bm{z}$. The Jacobian $J(\bm{X}(I) \to \bm{Z})$ of the linear transformation is a constant that doesn't depend on \(\bm{z}\) and therefore is dropped in the proportional sign.

Eq.\eqref{obj} is the computable-discretized posterior of EFiGP. In this paper, we consider the Maximum A Posteriori (MAP) as a fast point estimate from EFiGP, while the Posterior Mean and the Posterior Interval are the formal Bayesian inference results that further quantify the uncertainty.

\section{Simulation results}
In this section, we study the performance of EFiGP on three real-world systems: the FitzHugh-Nagumo (FN) \citep{Fitzhugh1961, Nagumo1962}, the Lotka-Volterra (LV) \citep{Lotka1932}, and the Hes1 system \citep{Hirata2002}. Since LV and Hes1 are strictly positive systems, we apply a log transformation to both. We then compare our method with the vanilla Bayesian GP method of MAGI, demonstrating that our proposed method improves the accuracy of inference results while significantly reducing run time.

\textbf{Data generation:} All ground truth data are simulated through numerical integration. Since FN, LV, and Hes1 are oscillators, we generate the true trajectories that cover approximately four to five cycles. To generate the noisy observations $\bm{y}(\bm{\tau})$, we use 41 equally spaced data points from the first half period as training, covering about two cycles, with added i.i.d. Gaussian random noise. The second half period is reserved for out of sample prediction evaluation. Thus, only 41 observations are available for each component. We also assume that all components are observed at the observation time points, and we use the same noise level for all components. 

\textbf{Benchmark models:} The MAGI framework \citep{yang2021a} has demonstrated better performance in previous comparisons with other Bayesian methods. In this study, we evaluate our proposed EFiGP approach against the state-of-the-art MAGI using varying discretization levels. For the inference process, we apply a much denser set of discretization points, increasing from the original 41 to 81, 161, 321, 641, and 1,281 points to show the effect on computational speed and accuracy.

\textbf{Evaluation Metric:} To evaluate model performance in recovering both the true parameters and system trajectories, we use the root mean squared error (RMSE) \citep{yang2021a}. For parameter estimation, we compute the absolute error between the inferred parameters and the pre-set true values. For each component’s trajectory, we focus on the period of observation, together with one extended period of the same length that does not have any observation. Given that the observation time points differ from the discretization time points, we compute the RMSE over 2,561 predetermined equally spaced time points along the reconstructed trajectories. The reconstructed trajectory is generated via numerical integration, using the inferred initial condition \(x_0\) (i.e., the first point of the inferred trajectory \(x(I)\)) and the inferred parameters. Notably, numerical integration is employed only for evaluation and forecasting in EFiGP and is not required for in-sample fitting.

\textbf{EFiGP Setting:} Regarding discretization, since only 41 observation points are available for inference, we use discretizations of 41, 81, 161, 321, 641, and 1,281 equally spaced time points (e.g., 161 for $\bm{I} = \{ t_1, t_2, \ldots, t_{161} \}$). These values represent 1, 2, 4, 8, 16, and 32 times the density of the observation time points. For truncation, we gradually increase the Fourier series $l$ and spectral decomposition terms $j$ (e.g., 11, 21, 41, 81, etc.) until results converge and stabilize. For the GP covariance function, we use the Matérn kernel with a degree of freedom of 2.01, ensuring that the kernel is twice differentiable.

\begin{table}[H]
\caption{Computational cost comparison between MAGI and EFiGP on different systems with different discretization levels, measured by running time (seconds), based on 100 repetitions.} \label{tab:run-time}
\label{CCost}
\centering
\fontsize{4pt}{4pt}\selectfont
\begin{tabular}{lccccccc}
\toprule
Discretization & \multicolumn{2}{c}{FN System} & \multicolumn{2}{c}{Hes1 System} & \multicolumn{2}{c}{LV System} \\
\cmidrule(lr){2-3} \cmidrule(lr){4-5} \cmidrule(lr){6-7}
& EFiGP & MAGI & EFiGP & MAGI & EFiGP & MAGI \\
\midrule
41   & 8.02$\pm$0.81  & 11.0$\pm$3.20   & 9.39$\pm$2.08   & 20.5$\pm$3.38  & 9.91$\pm$1.40   & 15.1$\pm$2.18  \\
81   & 7.93$\pm$0.67  & 11.2$\pm$2.22   & 10.4$\pm$2.99   & 20.8$\pm$2.51  & 8.11$\pm$0.59   & 13.3$\pm$2.16  \\
161  & 6.44$\pm$0.68  & 9.66$\pm$0.34   & 9.12$\pm$0.65   & 14.4$\pm$0.79  & 7.07$\pm$1.32   & 10.1$\pm$0.18   \\
321  & 7.04$\pm$0.55  & 11.7$\pm$0.54   & 9.80$\pm$0.75   & 17.8$\pm$2.75  & 6.84$\pm$0.39   & 11.8$\pm$0.31   \\
641  & 7.62$\pm$0.31  & 17.7$\pm$1.36   & 10.3$\pm$0.67   & 28.5$\pm$2.85  & 7.79$\pm$0.34   & 18.6$\pm$1.68   \\
1281 & 7.41$\pm$0.44  & 39.1$\pm$1.23   & 10.5$\pm$0.36   & 62.4$\pm$4.60  & 7.59$\pm$0.44   & 39.6$\pm$1.02   \\
\bottomrule
\end{tabular}
\end{table}

\begin{table}[H]
\caption{Mean and standard deviations of RMSE for MAGI and EFiGP for each component on the LV, FN, and Hes1 systems }
\label{CC_RMSE}
\begin{center}
\fontsize{4pt}{4pt}\selectfont
\setlength{\tabcolsep}{3pt}
\begin{sc}
\begin{tabular}{lcccccccc}
\toprule
System & Component & Method & 41 & 81 & 161 & 321 & 641 & 1281 \\
\midrule
\multirow{4}{*}{FN} 
    & \multirow{2}{*}{$x_1$} & EFiGP & 0.70$\pm$0.38 & 0.89$\pm$0.36 & 0.21$\pm$0.05 & 0.28$\pm$0.14 & 0.31$\pm$0.13 & 0.28$\pm$0.12 \\
    &                        & MAGI   & 0.42$\pm$0.28 & 0.48$\pm$0.21 & 0.29$\pm$0.11 & 0.30$\pm$0.13 & 0.39$\pm$0.15 & 0.43$\pm$0.15 \\
    & \multirow{2}{*}{$x_2$} & EFiGP & 0.26$\pm$0.16 & 0.34$\pm$0.16 & 0.09$\pm$0.04 & 0.10$\pm$0.04 & 0.11$\pm$0.04 & 0.09$\pm$0.04 \\
    &                        & MAGI   & 0.17$\pm$0.12 & 0.26$\pm$0.09 & 0.22$\pm$0.06 & 0.20$\pm$0.04 & 0.20$\pm$0.05 & 0.21$\pm$0.04 \\
\midrule
\multirow{6}{*}{Hes1} 
    & \multirow{2}{*}{$\log(x_1)$} & EFiGP & 0.32$\pm$0.15 & 0.24$\pm$0.09 & 0.19$\pm$0.06 & 0.17$\pm$0.04 & 0.12$\pm$0.03 & 0.09$\pm$0.02 \\
    &                        & MAGI   & 0.30$\pm$0.16 & 0.22$\pm$0.09 & 0.21$\pm$0.12 & na & na & na \\
    & \multirow{2}{*}{$\log(x_2)$} & EFiGP & 0.23$\pm$0.12 & 0.17$\pm$0.08 & 0.11$\pm$0.04 & 0.07$\pm$0.02 & 0.09$\pm$0.02 & 0.11$\pm$0.02 \\
    &                        & MAGI   & 0.22$\pm$0.12 & 0.15$\pm$0.07 & 0.12$\pm$0.07 & na & na & na \\
    & \multirow{2}{*}{$\log(x_3)$} & EFiGP & 0.64$\pm$0.28 & 0.47$\pm$0.17 & 0.37$\pm$0.13 & 0.34$\pm$0.08 & 0.21$\pm$0.05 & 0.18$\pm$0.05 \\
    &                        & MAGI   & 0.59$\pm$0.29 & 0.43$\pm$0.18 & 0.38$\pm$0.19 & na & na & na \\
\midrule
\multirow{4}{*}{LV}
    & \multirow{2}{*}{$\log(x_1)$} & EFiGP & 0.16$\pm$0.04 & 0.13$\pm$0.07 & 0.10$\pm$0.06 & 0.06$\pm$0.03 & 0.04$\pm$0.03 & 0.06$\pm$0.02 \\
    &                        & MAGI   & 0.17$\pm$0.12 & 0.12$\pm$0.09 & 0.09$\pm$0.05 & 0.06$\pm$0.03 & 0.06$\pm$0.03 & 0.11$\pm$0.05 \\
    & \multirow{2}{*}{$\log(x_2)$} & EFiGP & 0.23$\pm$0.06 & 0.18$\pm$0.10 & 0.15$\pm$0.08 & 0.08$\pm$0.04 & 0.05$\pm$0.03 & 0.06$\pm$0.02 \\
    &                        & MAGI   & 0.25$\pm$0.18 & 0.18$\pm$0.14 & 0.12$\pm$0.09 & 0.08$\pm$0.05 & 0.06$\pm$0.04 & 0.11$\pm$0.07 \\
\bottomrule
\end{tabular}
\end{sc}
\end{center}
\end{table}

\subsection{FN system}
\begin{figure}[H]
\centering
\includegraphics[width=6cm]{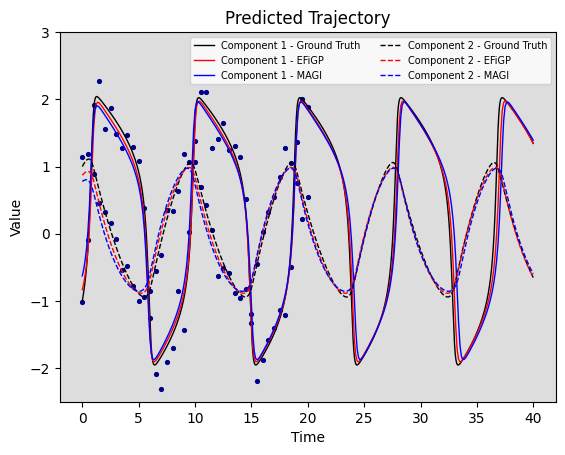}
\caption{Predicted trajectory from EFiGP (red solid and dashed line) and from MAGI (blue solid and dashed line) for a 1281 discretization size on the FN system with ground-truth trajectory (black solid and dashed line) and 41 observed data points.}
\label{FNpred}
\end{figure}

The FitzHugh-Nagumo (FN) system was introduced by \citet{Fitzhugh1961,Nagumo1962} to model the activation of excitable systems such as neurons. It is a two-component system governed by the following ODEs:

\begin{equation}
\begin{cases}
\dot{x}_1 = c \left( x_1 - \frac{x_1^3}{3} + x_2 \right), \\
\dot{x}_2 = -\frac{x_1 - a + bx_2}{c},
\end{cases}
\end{equation}

where \( a = 0.2 \), \( b = 0.2 \), \( c = 3 \), and \( x(0) = (-1, 1) \) are the true parameter values and initial conditions. We simulated 100 datasets with a noise level of 0.2 across both components. Fig.\ref{FNpred} visualizes one example dataset and evaluation period by using EFiGP and MAGI. 

Tab.\ref{tab:run-time} shows that the average runtime of EFiGP no longer increases with discretization sizes. At a discretization size of 1281, EFiGP is approximately six times faster than MAGI. The stabilized truncation number is shown in Tab.\ref{tab:fn-truncation-number} in SI, which plateaued at 81 Eigenvalues and 41 Fourier series after discretization gets to 321.

In terms of performance, EFiGP consistently yields more accurate results across the two components as the discretization increased, along with improved parameter estimation accuracy compared to MAGI. EFiGP has the most outperformance when the discretization is dense enough at 161. As seen in Tab.~\ref{CC_RMSE} and Tab.~\ref{fpep}, EFiGP yields more stable results for each component and improves parameter estimation accuracy as the discretization increases beyond 161 (four times denser). Specifically, the EFiGP stabilizes at 161, and further increasing the discretization size beyond 321 does not further improve or degrade the results. On the contrary, the MAGI results on $x_1$ deteriorate as the discretization increases beyond 321.

\begin{table}[H]
\caption{Mean and standard deviations of Absolute Error for MAGI and EFiGP for each parameter on the FN system}
\label{fpep}
\begin{center}
\fontsize{5pt}{5pt}\selectfont
\setlength{\tabcolsep}{3pt}
\begin{sc}
\begin{tabular}{lccccccc}
\toprule
 & & \textbf{41} & \textbf{81} & \textbf{161} & \textbf{321} & \textbf{641} & \textbf{1281}  \\
\midrule
\multirow{3}{*}{\textbf{EFiGP}} 
    & \textnormal{a}  & \hfill .020$\pm$.013 & \hfill .025$\pm$.022 & \hfill .027$\pm$.026 & \hfill .030$\pm$.026 & \hfill .031$\pm$.025 & \hfill .031$\pm$.024  \\
    & \textnormal{b}  & \hfill .176$\pm$.116 & \hfill .121$\pm$.099 & \hfill .190$\pm$.111 & \hfill .257$\pm$.114 & \hfill .264$\pm$.112 & \hfill .233$\pm$.103  \\
    & \textnormal{c}  & \hfill .100$\pm$.075 & \hfill .174$\pm$.094 & \hfill .068$\pm$.047 & \hfill .075$\pm$.051 & \hfill .067$\pm$.046 & \hfill .050$\pm$.034  \\
\midrule
\multirow{3}{*}{\textbf{MAGI}}   
    & \textnormal{a}  & \hfill .026$\pm$.019 & \hfill .017$\pm$.014 & \hfill .028$\pm$.017 & \hfill .033$\pm$.017 & \hfill .032$\pm$.017 & \hfill .031$\pm$.018  \\
    & \textnormal{b}  & \hfill .120$\pm$.082 & \hfill .229$\pm$.141 & \hfill .388$\pm$.144 & \hfill .475$\pm$.123 & \hfill .500$\pm$.099 & \hfill .500$\pm$.085  \\
    & \textnormal{c}  & \hfill .126$\pm$.083 & \hfill .241$\pm$.105 & \hfill .288$\pm$.104 & \hfill .277$\pm$.097 & \hfill .251$\pm$.087 & \hfill .231$\pm$.080  \\
\bottomrule
\end{tabular}
\end{sc}
\end{center}
\end{table}

\subsection{Hes1 system}

\begin{figure}[H]
\centering
\includegraphics[width=6cm]{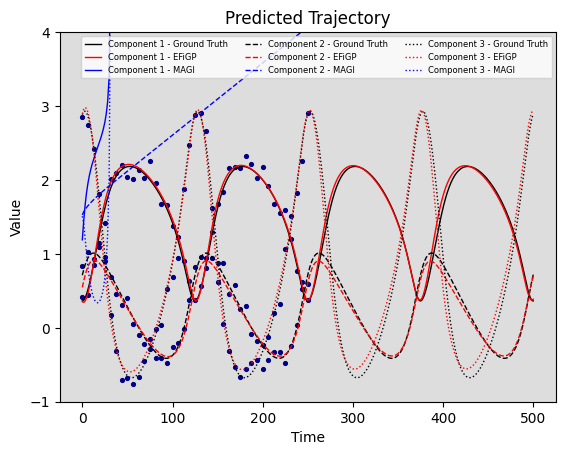}
\caption{Predicted trajectory from EFiGP (red solid, dashed and dotted line) and from MAGI (blue solid, dashed and dotted line) for a 1281 discretization size on the log-transformed Hes1 system with ground-truth trajectory (black solid, dashed and dotted line) and 41 observed data points. At 1281 discretization, MAGI failed to converge while EFiGP still produce meaningful results. }
\label{Hespred}
\end{figure}

The Hes1 system was introduced by \citet{Hirata2002} to model the oscillatory dynamics of the Hes1 protein level ($x_1$) and Hes1 mRNA level ($x_2$) under the influence of a Hes1-interacting factor ($x_3$). It is a three-component system governed by the following ODEs:

\begin{equation}
\begin{cases}
\dot{x}_1 = -ax_1x_3 + bx_2 - cx_1, \\
\dot{x}_2 = -dx_2 + \frac{e}{1 + x_1^2}, \\
\dot{x}_3 = -ax_1x_3 + \frac{f}{1 + x_1^2} - gx_3,
\end{cases}
\end{equation}

where the true parameter values are $a = 0.022$, $b = 0.3$, $c = 0.031$, $d = 0.028$, $e = 0.5$, $f = 20$, and $g = 0.3$. The initial condition is $x(0) = (1.438575, 2.037488, 17.90385)$. These parameter values and initial conditions are used to generate the ground-truth trajectory. We simulated 100 datasets with a log-normal noise of 0.1, using 41 observations. Fig.\ref{Hespred} visualizes one example dataset in log scale together with reconstructed trajectories on the fitting period and prediction period by using EFiGP and MAGI. 

\begin{table}[H]
\caption{Mean and standard deviations of Absolute Error for MAGI and EFiGP for each parameter on the Hes1 system}
\label{Hpep}
\begin{center}
\fontsize{4pt}{4pt}\selectfont
\setlength{\tabcolsep}{3pt}
\begin{sc}
\begin{tabular}{lccccccc}
\toprule
 & &\textbf{41} & \textbf{81} & \textbf{161} & \textbf{321} & \textbf{641} & \textbf{1281}  \\
\midrule
\multirow{7}{*}{\textbf{EFiGP}} & \textnormal{a} & \hfill 0.002$\pm$0.001& \hfill0.001$\pm$0.001 & \hfill 0.001$\pm$0.001 & \hfill 0.001$\pm$0.001 & \hfill 0.001$\pm$0.001 & \hfill 0.001$\pm$0.001  \\
                        & \textnormal{b} & \hfill 0.024$\pm$0.018 &  \hfill 0.023$\pm$0.017 & \hfill 0.022$\pm$0.019 & \hfill 0.039$\pm$0.031 & \hfill 0.059$\pm$0.039 & \hfill 0.071$\pm$0.043  \\
                        & \textnormal{c} & \hfill 0.004$\pm$0.003& \hfill 0.004$\pm$0.002 & \hfill 0.003$\pm$0.002 & \hfill 0.004$\pm$0.003 & \hfill 0.007$\pm$0.004 & \hfill 0.008$\pm$0.005  \\
                        & \textnormal{d} & \hfill 0.001$\pm$0.001 & \hfill 0.001$\pm$0.001  & \hfill 0.001$\pm$0.001 & \hfill 0.003$\pm$0.002 & \hfill 0.005$\pm$0.002 & \hfill 0.005$\pm$0.002  \\
                        & \textnormal{e} & \hfill 0.026$\pm$0.030& \hfill 0.024$\pm$0.016 & \hfill 0.026$\pm$0.018 & \hfill 0.035$\pm$0.028 & \hfill 0.082$\pm$0.041 & \hfill 0.112$\pm$0.045  \\
                        & \textnormal{f} & \hfill 10.156$\pm$0.092 & \hfill10.254$\pm$0.079  & \hfill 10.279$\pm$0.085 & \hfill 10.296$\pm$0.089 & \hfill 10.304$\pm$0.089 & \hfill 10.315$\pm$0.086  \\
                        & \textnormal{g}& \hfill 0.194$\pm$0.019& \hfill 0.191$\pm$0.019 & \hfill 0.167$\pm$0.022 & \hfill 0.153$\pm$0.023 & \hfill 0.152$\pm$0.023 & \hfill 0.151$\pm$0.023  \\

\midrule
\multirow{7}{*}{\textbf{MAGI}}   & \textnormal{a}  & \hfil 0.002$\pm$0.001& \hfill0.001$\pm$0.001 & \hfill 0.002$\pm$0.003 &  na &  na &  na  \\
                        & \textnormal{b} & \hfill0.028$\pm$0.020 & \hfill0.023$\pm$0.017 & \hfill 0.028$\pm$0.045 &  na &  na &  na  \\
                        & \textnormal{c} & \hfill 0.004$\pm$0.003& \hfill0.004$\pm$0.002 & \hfill 0.004$\pm$0.005 &  na &  na &  na  \\
                        & \textnormal{d} & \hfill0.001$\pm$0.001& \hfill 0.001$\pm$0.001& \hfill 0.001$\pm$0.002 &  na &  na &  na  \\
                        & \textnormal{e} & \hfill 0.026$\pm$0.017 & \hfill0.025$\pm$0.016 & \hfill 0.037$\pm$0.055 &  na &  na &  na  \\
                        & \textnormal{f} & \hfill 10.142$\pm$0.099& \hfill 10.247$\pm$0.081& \hfill 10.279$\pm$0.089 & na &  na &  na  \\
                        & \textnormal{g} & \hfill0.195$\pm$0.018& \hfill 0.189$\pm$0.019& \hfill 0.167$\pm$0.024 &  na &  na &  na  \\
\bottomrule
\end{tabular}
\end{sc}
\end{center}
\end{table}

The run time comparison is shown in Tab.\ref{tab:run-time}.
The average runtime of EFiGP for this system over 100 repetitions remains constant across all discretization sizes. Notably, after a discretization size of 161, EFiGP becomes twice as fast as MAGI. For a discretization size of 641, EFiGP is about four times faster, and at 1281, it is approximately six times faster. In contrast, the runtime of MAGI increases almost linearly as the discretization becomes denser. The stabilized truncation number is shown in Tab.\ref{tab:hes1-truncation-number} in SI, which again plateaued at 81 Eigenvalues and 41 Fourier series after discretization gets to 321.

Tab.\ref{CC_RMSE} summarizes the accuracy of the reconstructed trajectories for the three system components, while Tab.\ref{Hpep} reports the estimation accuracy of the parameters. Both tables present results across different discretization levels. In general, EFiGP demonstrates improved performance in trajectory reconstruction as the discretization level increases up to 641. Meanwhile, parameter estimation accuracy stabilizes at a lower discretization level of 161. This may be attributed to better recovery of weakly identifiable parameter combinations that deviate from the true values but yield similar trajectories. In contrast, MAGI fails to converge when the discretization exceeds 321, underscoring the enhanced robustness of EFiGP.

\subsection{LV system}

\begin{figure}[H]
\centering
\includegraphics[width=6cm]{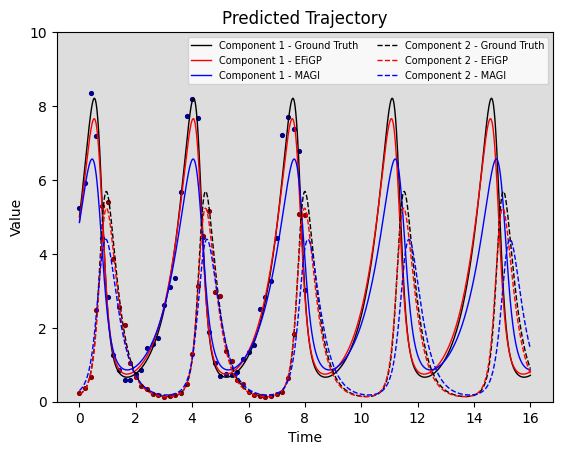}
\caption{Predicted trajectory from EFiGP (red solid and dashed line) and from MAGI (blue solid and dashed line) for a 1281 discretization size on the LV system with ground-truth trajectory (black solid and dashed line) and 41 observed data points.}
\label{Lvpred}
\end{figure}

The Lotka-Volterra (LV) system was introduced by \citet{Lotka1932} to model the dynamics of predator-prey interactions. It is a two-component system governed by the following ODEs:

\begin{equation}
\begin{cases}
\dot{x}_1 = ax_1 - bx_1x_2, \\
\dot{x}_2 = cx_1x_2 - dx_2,
\end{cases}
\end{equation}

where \( a = 1.5 \), \( b = 1 \), \( c = 1 \), and \( d = 3 \) are the true parameter values, and \( x(0) = (5, 0.2) \) is the initial condition. These parameters are used to generate the ground-truth trajectory. We simulated 100 datasets with a log-normal noise of 0.1, using 41 observations. Fig.\ref{Lvpred} visualizes one example dataset in exponential scale together with reconstructed trajectories on the fitting period and prediction period by using EFiGP and MAGI. 

Tab.~\ref{CCost} shows the run time comparison between the MAGI and EFiGP. The average runtime of EFiGP no longer increases with discretization sizes. In contrast, the runtime of MAGI grows substantially with finer discretization. Importantly, after a discretization size of 321, EFiGP becomes twice as fast as MAGI. At a discretization size of 1281, EFiGP is approximately 5 times faster than MAGI. The stabilized truncation number is shown in Tab.\ref{tab:lv-truncation-number} in SI, which plateaued at 81 Eigenvalues and 41 Fourier series at an even earlier stage of discretization level 161.

\begin{figure}[H]
    \centering
    \subfigure{%
        \includegraphics[width=6cm]{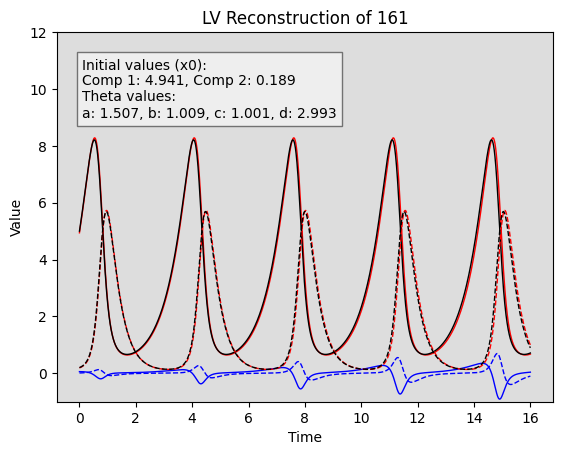}
    }
    \hspace{0.1cm} % Adjust the space between the two plots
    \subfigure
    {%
        \includegraphics[width=6cm]{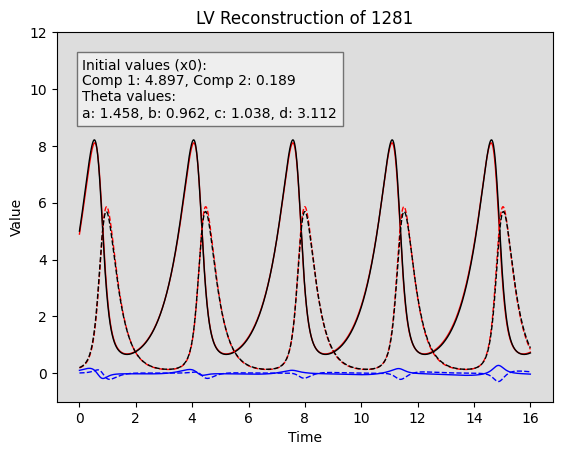}
    }
\caption{Single-dataset comparison between the predicted trajectory (red solid and dashed lines) and the ground-truth trajectory (black solid and dashed lines) at a discretization level of 161 (LHS) and 1,281 (RHS). Also shown is a different trajectory (blue solid and dashed lines) inferred using the estimated initial values and parameters. This comparison is conducted on a single dataset for illustrative purposes. The true parameter values are \( a = 1.5 \), \( b = 1 \), \( c = 1 \), and \( d = 3 \), with initial conditions \( x_1(0) = 5 \) and \( x_2(0) = 0.2 \). As evident from the figure, at a discretization level of 161 (LHS), the inferred parameters are more accurate. However, at a finer discretization level of 1,281 (RHS), the trajectory RMSE is lower despite greater parameter estimation errors. This highlights the phenomenon of weakly identifiable parameters, where a parameter set with higher error can still yield trajectories with improved accuracy.}
\label{error}
\end{figure}

Tab.~\ref{CC_RMSE} and Tab.~\ref{LV_RMSE} present the performance of EFiGP and MAGI across varying discretization levels. EFiGP consistently demonstrates more stable trajectory reconstruction across all components, with accuracy improving as the discretization increases beyond 321 (eight times denser). Notably, EFiGP outperforms MAGI in trajectory accuracy for components \(x_1\) and \(x_2\) at higher discretization levels. However, at finer discretization levels beyond 161, EFiGP's parameter estimation accuracy deteriorates, despite continued improvements in trajectory reconstruction. Further analysis reveals that due to issues with differentiability, parameter combinations that deviate more from the ground truth can produce trajectories nearly indistinguishable from the actual ones, as shown in SI Fig.\ref{error}.

\begin{table}[H]
\caption{Mean and standard deviations of Absolute Error for MAGI and EFiGP for each parameter on the LV system with a tuned learning rate}
\label{LV_RMSE}
\setlength{\tabcolsep}{3pt}
\begin{center}
\fontsize{5pt}{5pt}\selectfont
\begin{sc}
\begin{tabular}{lccccccc}
\toprule
 & & \textbf{41}& \textbf{81} & \textbf{161} & \textbf{321} & \textbf{641} & \textbf{1281}  \\
\midrule
\multirow{6}{*}{\textbf{EFiGP}} 
    & \textnormal{a} & \hfill .026$\pm$.019  & \hfill .026$\pm$.019 & \hfill .036$\pm$.025 & \hfill .054$\pm$.028 & \hfill .067$\pm$.028 & \hfill .078$\pm$.024   \\
    & \textnormal{b} & \hfill .027$\pm$.019 & \hfill .027$\pm$.020 & \hfill .033$\pm$.025 & \hfill .049$\pm$.029 & \hfill .056$\pm$.028 & \hfill .040$\pm$.026   \\
    & \textnormal{c} & \hfill .028$\pm$.019 & \hfill .029$\pm$.019 & \hfill .036$\pm$.025 & \hfill .057$\pm$.028 & \hfill .067$\pm$.027 & \hfill .073$\pm$.019   \\
    & \textnormal{d} & \hfill .051$\pm$.032 & \hfill .049$\pm$.034 & \hfill .066$\pm$.044 & \hfill .102$\pm$.053 & \hfill .121$\pm$.055 & \hfill .159$\pm$.034   \\
    % & \textnormal{$x_1(0)$} & \hfill .043$\pm$.023 & \hfill .028$\pm$.017 & \hfill .026$\pm$.019 & \hfill .035$\pm$.017 & \hfill  .040$\pm$.015 & \hfill .038$\pm$.015 \\
    % & \textnormal{$x_2(0)$} & .062$\pm$.042 & .051$\pm$.036 & .032$\pm$.027 &.022$\pm$.017 & .025$\pm$.014 & .038$\pm$.028 \\
\midrule
\multirow{4}{*}{\textbf{MAGI}} 
    & \textnormal{a} & \hfill .028$\pm$.023 & \hfill .026$\pm$.021 & \hfill .035$\pm$.024 & \hfill .056$\pm$.028 & \hfill .079$\pm$.029 & \hfill .119$\pm$.035  \\
    & \textnormal{b} & \hfill .028$\pm$.022 & \hfill .027$\pm$.021 & \hfill .031$\pm$.022 & \hfill .049$\pm$.027 & \hfill .058$\pm$.029 & \hfill .042$\pm$.029  \\
    & \textnormal{c} & \hfill .022$\pm$.019 & \hfill .022$\pm$.018 & \hfill .032$\pm$.021 & \hfill .058$\pm$.025 & \hfill .069$\pm$.025 & \hfill .072$\pm$.022  \\
    & \textnormal{d} & \hfill .045$\pm$.036 & \hfill .044$\pm$.035 & \hfill .059$\pm$.042 & \hfill .105$\pm$.052 & \hfill .139$\pm$.053  & \hfill .170$\pm$.050  \\
    % & \textnormal{$x_1(0)$} & .042$\pm$.034 & .027$\pm$.022 & .023$\pm$.022 & .028$\pm$.022 & .032$\pm$.019 & .028$\pm$.022 \\
    % & \textnormal{$x_2(0)$} & .081$\pm$.049 & .072$\pm$.043 & .057$\pm$.036 & .043$\pm$.028 & .040$\pm$.032 & .080$\pm$.054 \\
\bottomrule
\end{tabular}
\end{sc}
\end{center}
\end{table}

\section{Discussion and Conclusion}

In this paper, we introduce a methodology for inferring dynamical systems using eigen-decomposed, Fourier-transformed, and physics-informed Gaussian Process. The Fourier transform provides several key advantages over working in the original space: (1) Incorporating physics information in the Fourier domain averages discrepancies in derivative information between the ODEs and the GP across the entire domain, rather than limiting it to discretization points. (2) Adding more frequency terms progressively introduces orthogonal information, while increasing the number of discretization points often leads to diminishing returns due to growing correlations. (3) For oscillatory ODEs, enforcing physics information in the Fourier domain ensures long-term reliability, whereas discretization points may fail to generalize beyond their coverage. Additionally, our method achieves better computational efficiency and accuracy as the density of discretization points increases. It outperforms existing GP-based approaches in inference accuracy on benchmark examples, with significantly faster computation times.

The primary parameter requiring tuning in our approach is the truncation number for both the eigen-decomposition and the Fourier series, as well as the discretization number inherited from previous methods. In practice, we recommend gradually increasing these numbers until the results stabilize, following the guidance of \citet{yang2021a} for setting the discretization number. Specifically, stabilization is achieved when further increases in the truncation number no longer affect accuracy or convergence. For each system analyzed, the truncation numbers used for spectral decomposition and the Fourier series are summarized in \S~\ref{Trun}. As we increased the discretization of the interpolated data -- from the original 41 observations to denser grids up to 1,281 points -- we consistently observed stabilization at \( j = 81 \) for the eigen-decomposition and \( l = 41 \) for the Fourier series. These values represent the optimal trade-off between computational efficiency and accuracy for the systems under study.

Fourier transformation and spectral decomposition have demonstrated significant success in achieving accurate approximations, particularly for system reconstruction, which remains stable even as discretization becomes denser. However, parameter estimation accuracy still degrades with increased discretization, largely due to the weak identifiability inherent in ODE systems. Addressing this limitation is an important direction for future research. Moreover, while this study focuses on point estimation, future work should explore uncertainty quantification within the Bayesian framework. This would enhance the robustness of our method and enable recovery of the full range of plausible weakly identifiable parameters.

\section*{Impact Statement}
Our work leverages Fourier transformation and spectral decomposition techniques to enhance physics-informed Gaussian Process, offering a computationally efficient framework for parameter inference and system reconstruction in ODEs inverse problems using observational data. This approach establishes a foundation for robust applications across diverse fields, including biology, engineering, and physics, where understanding the interaction between parameters and trajectories is essential. Our research has the potential to significantly improve the reliability, efficiency, and interpretability of data-driven models for complex dynamical systems.

\bibliographystyle{plainnat}

\bibliography{Reference}

\section{supplementary materials}

\subsection{GP Smoothing}
\label{GPSm}
Gaussian Process (GP) smoothing is a powerful non-parametric technique used to model and predict complex, noisy data. Formally, a Gaussian Process is specified by its mean function \( m(\bm{\tau}) \), which is often assumed to be zero for simplicity, and its covariance function \( k(\bm{\tau}, \bm{\tau'}) \), which defines the relationship between different points in the input space. For any finite set of points \( \bm{\tau} = (\tau_1, \tau_2, \ldots, \tau_N) \), we have:

\[
\bm{X}(\bm{\tau}) \sim \mathcal{GP}(m(\bm{\tau}),  k(\bm{\tau}, \bm{\tau'})),
\]

Given a set of observed data points \(\{(\tau_i, y_i)\}_{i=1}^N\), where \( \bm{Y}(\bm{\tau}) = \bm{X}(\bm{\tau}) + \bm{\epsilon} \) and \( \epsilon_i \sim \mathcal{N}(0, \sigma^2) \) represents noise, GP smoothing aims to infer the function \( \bm{X}(\bm{\tau}) \).

The joint distribution of the observed outputs \( \bm{Y}(\bm{\tau}) \) and the predicted values \( \bm{X}(\bm{\tilde{\tau}}) \) at test points \( \bm{\tilde{\tau}} = (\tilde{\tau}_1, \tilde{\tau}_2, \ldots, \tilde{\tau}_m) \) is given by:

\[
\begin{bmatrix}
\bm{Y}(\bm{\tau}) \\
\bm{X}(\bm{\tilde{\tau}})
\end{bmatrix}
\sim \mathcal{N} \left( 0, 
\begin{bmatrix}
K(\bm{\tau}, \bm{\tau}) + \sigma^2 I & K(\bm{\tau}, \bm{\tilde{\tau}}) \\
K(\bm{\tilde{\tau}}, \bm{\tau}) & K(\bm{\tilde{\tau}}, \bm{\tilde{\tau}})
\end{bmatrix}
\right),
\]

where \( K(\bm{\tau}, \bm{\tau}) \) is the covariance matrix evaluated at the training points, \( K(\bm{\tau}, \bm{\tilde{\tau}}) \) is the covariance between the training and test points, and \( K(\bm{\tilde{\tau}}, \bm{\tilde{\tau}}) \) is the covariance matrix at the test points.

The posterior distribution over the function values at the test points \( \bm{X}(\bm{\tilde{\tau}}) \), given the observed data, is:

\[
\bm{X}(\bm{\tilde{\tau}}) \mid \bm{\tau}, \bm{Y}(\bm{\tau}), \bm{\tilde{\tau}} \sim \mathcal{N}(\tilde{\mu}, \tilde{\text{cov}}),
\]

where the mean and covariance of the posterior distribution are given by:

\[
\tilde{\mu} = K(\bm{\tilde{\tau}}, \bm{\tau}) \left[K(\bm{\tau}, \bm{\tau}) + \sigma^2 I\right]^{-1} \bm{Y}(\bm{\tau}),
\]

\[
\tilde{\text{cov}} = K(\bm{\tilde{\tau}}, \bm{\tilde{\tau}}) - K(\bm{\tilde{\tau}}, \bm{\tau}) \left[K(\bm{\tau}, \bm{\tau}) + \sigma^2 I\right]^{-1} K(\bm{\tau}, \bm{\tilde{\tau}}).
\]

GP smoothing provides not only predictions but also measures of uncertainty, making it a robust method for modeling and interpreting noisy data. Its applications span various fields, including geostatistics, machine learning, and time-series analysis.

\subsection{Full $d$ notation}
\label{fulld}
\textbf{Prior:} We impose a general prior \(\pi(\cdot)\) on \(\theta\) and an independent GP prior on each component \(\bm{X}_d(t)\):
\begin{equation}
    \bm{X}_d(t) \sim \mathcal{GP}(\bm{\mu}_d, \bm{\mathcal{K}}_d) \quad t \in [0, T]
\end{equation}
where the mean function \(\bm{\mu}_d : \mathbb{R} \to \mathbb{R}\) and the positive-definite covariance function \(\bm{\mathcal{K}}_d : \mathbb{R} \times \mathbb{R} \to \mathbb{R}\) are parameterized by hyperparameters \(\bm{\phi}_d\).

\textbf{Likelihood:} For any finite set of time points \(\bm{\tau}_d\), \(\bm{X}_d(\bm{\tau}_d)\) has a multivariate Gaussian distribution:
\begin{equation}
    \bm{Y}_d(\bm{\tau}_d) \mid \bm{X}_d(\bm{\tau}_d) = \bm{x}_d(\bm{\tau}_d) \sim \mathcal{N}(\bm{x}_d(\bm{\tau}_d), \sigma_d^2 \bm{I}_{N_d})
\end{equation}

We define the random variable \(W\) quantifying the difference between the time derivative \(\dot{\bm{X}}_d(t)\) of the GP and the ODE structure:
\begin{equation}
    W = \sup_{d = 1, \ldots, D; t \in [0, T]} \left|{\dot{\bm{X}}_d(t)} - f(\bm{X}_d(t), \theta, t) \right|
\label{Mcons}
\end{equation}
Since \(W\) cannot be computed analytically, we approximate it with \(W_{I}\) using a finite discretization:
\begin{equation}
    W_{I} = \max_{d = 1, \ldots, D; t \in I} \left|{\dot{\bm{X}}_d(t)} - f(\bm{X}_d(t), \theta, t) \right|
\end{equation}

\textbf{Posterior:} The computable posterior distribution is:
\begin{equation}
    p_{\bm\Theta, \bm{X}_d(I) \mid W_{I,d}, \bm{Y}_d(\bm{\tau}_d)}(\theta, \bm{x}_d(I) \mid W_{I,d} = 0, \bm{Y}_d(\bm{\tau}_d) = \bm{y}_d(\bm{\tau}_d))
\end{equation}
By Bayes’ rule, we have:
\begin{equation}
\begin{aligned}
    p_{\Theta, \bm{X}_d(I) \mid W_{I,d}, \bm{Y}_d(\bm{\tau}_d)}(\theta, \bm{x}_d(I) \mid W_{I,d} = 0, \bm{Y}_d(\bm{\tau}_d) = \bm{y}_d(\bm{\tau}_d)) &\propto \\
    P(\Theta = \theta, \bm{X}_d(I) = \bm{x}_d(I), W_{I,d} = 0, \bm{Y}_d(\bm{\tau}_d) = \bm{y}_d(\bm{\tau}_d))
\end{aligned}
\end{equation}
The closed form of the right-hand side is:
\begin{equation}
\begin{split}
    &P(\Theta = \theta, \bm{X}_d(I) = \bm{x}_d(I), W_{I,d} = 0, \bm{Y}_d(\bm{\tau}_d) = \bm{y}_d(\bm{\tau}_d)) \\
    &= \pi_{\Theta}(\theta) \times P(\bm{X}_d(I) = \bm{x}_d(I) \mid \Theta = \theta) \\
    &\times P(\bm{Y}_d(\bm{\tau}_d) = \bm{y}_d(\bm{\tau}_d) \mid \bm{X}_d(I) = \bm{x}_d(I), \Theta = \theta) \\
    &\times P(W_{I,d} = 0 \mid \bm{Y}_d(\bm{\tau}_d) = \bm{y}_d(\bm{\tau}_d), \bm{X}_d(I) = \bm{x}_d(I), \Theta = \theta)
\end{split}
\label{MAGI_d}
\end{equation}

The ODE information part of EFiGP is

\begin{equation}
W_I^\mathcal{F} = \max_{d = 1,\ldots,D; t\in I} \left|\mathcal{F}[\dot{\bm{X}}_d(t)] - \mathcal{F}[f(\bm{X}_d(t),\theta,t)] \right|
\end{equation}

Where the set \(I = (t_1, t_2, \ldots, t_n)\) with \(n\) discretization points. Also, we can easily obtain the computational form by Lemma~\ref{DFT} since \(\dot{\bm{X}}_d\) is a joint Gaussian distribution.

Secondly, since posterior sampling or maximum a posteriori (MAP) optimization on \(\bm{X}_d(I)\) in the objective function (\ref{MAGI}) incurs a high cost when the set becomes denser, we propose an efficient way to handle \(\bm{X}_d(I)\) by using spectral decomposition (Lemma~\ref{Eigen}). We consider the change of variable (orthogonally reparametrize) \(\bm{X}_d(I)\) to \(\bm{z}_d = (z_{d1}, \ldots, z_{dn})\), using the matrix square root from the spectral decomposition of the prior variance and covariance matrix:
\begin{equation}
\bm{X}_d(I) = \bm{\mu}_d(I) + \bm{V}_{d(j)}\bm{\Lambda}_{d(j)}^{\frac{1}{2}}\bm{z}_d = \bm{\mu}_d(I) + \sum_{i=1}^{j} z_{di} \sqrt{\lambda_{di}} \bm{v}_{di}
\end{equation}

where \(\lambda_{di}, \bm{v}_{di}\) are eigenvalues and eigenvectors of \(\bm{\mathcal{K}}_d(I, I)\). By this form, we can also truncate the summation and keep the first \(j\) terms to save computational cost over all parts of the objective function (\ref{MAGI}).

\textbf{Objective Function of EFiGP with full $d$:}

\begin{equation}
\begin{split}
& P(\Theta = \theta, \bm{Z}_d = \bm{z}_d \mid W^\mathcal{F}_{I,d} = 0, \bm{Y}_d(\tau_d) = \bm{y}_d(\tau_d))  \\
\propto  & P(\Theta = \theta, \bm{X}_d(I) = \bm{x}_d(I), W^\mathcal{F}_{I,d} = 0, \bm{Y}_d(\tau_d) = \bm{y}_d(\tau_d))\times J(\bm{X}_d(I) \to \bm{Z}_d)\\
\propto &\ \pi_{\Theta}(\theta) \exp \{ -\frac{1}{2} \left[ |I| \log(2\pi) + \bm{z}_d^T\bm{z}_d \right. \\
&+ |I| \log(2\pi) + \log| \bm{K}_{d,k}^{\mathcal{F}}| + \left\|\mathcal{F}[f^{\theta,\bm{x}}_{d,I}]_k - \mathcal{F}[m_d\{\bm{V}_{d(j)}\bm{\Lambda}_{d(j)}^{\frac{1}{2}}\bm{z}_d\} ]_k\right\|^2_{(\bm{K}_{d,k}^{\mathcal{F}})^{-1}} \\
&+ N_d \log(2\pi\sigma_d^2) + \| \bm{V}_{d(j)}\bm{\Lambda}_{d(j)}^{\frac{1}{2}}\bm{z}_d(\tau_d) - \bm{y}_d(\tau_d) \|^2_{\sigma_d^{-2}} \left. \right] \}
\end{split}
\end{equation}

where \(\bm{V}_{d(j)} = (\bm{v}_{d1}, \ldots, \bm{v}_{dj})\), \(\bm{\Lambda}_{d(j)} = \mathrm{diag}(\lambda_{d1}, \ldots, \lambda_{dj})\), and \(\bm{K}_{d,k}^{\mathcal{F}}\) can be obtained by property with the truncated number \(k \in \mathbb{N}\). The (integral) Jacobian of the linear transformation is a constant that doesn't depend on \(\bm{z}_d\) and therefore is dropped in the proportional sign.

Also, the short notations are:
\begin{align*}
    & \|\bm{v}_d\|^2_A = \bm{v}_d^T A \bm{v}_d \\
    & m_d ='\bm{\mathcal{K}}_d(I,I)\bm{\mathcal{K}}_d(I,I)^{-1} \\
    & K_d = \bm{\mathcal{K}}_d''(I,I)-'\bm{\mathcal{K}}_d(I,I)\bm{\mathcal{K}}_d(I,I)^{-1}\bm{\mathcal{K}}_d'(I,I)
\end{align*}
where \( '\bm{\mathcal{K}}_d = \frac{\partial}{\partial{s}}\bm{\mathcal{K}}_d(s,t)\), \(\bm{\mathcal{K}}_d' = \frac{\partial}{\partial{t}}\bm{\mathcal{K}}_d(s,t)\), and \(\bm{\mathcal{K}}_d'' = \frac{\partial^2}{\partial{s}\partial{t}}\bm{\mathcal{K}}_d(s,t)\).

After optimizing \(\bm{z}_d\), we can transfer it back to the original space with a dense discretization by:
\begin{equation}
\bm{X}_d(I) = \bm{\mu}_d(I) + \bm{V}_{d(j)}\bm{\Lambda}_{d(j)}^{\frac{1}{2}}\bm{z}_d
\end{equation}

\subsection{Closed form of Transformation Matrix}
\label{transma}
Given $\bm{X} \sim \mathcal{N}(0,\bm{\Sigma})$, where $\bm{X}\in \mathbb{R}^n$ and $\bm{\Sigma}\in\mathbb{R}^{n\times n}$, and a Discrete Fourier Transform (DFT) matrix, $\bm{A}$, we can write $\bm{Y}\in \mathbb{C}^n$ as:
\[
\bm{Y} = \bm{A}\bm{X} = \bm{U} + i\bm{V},
\]
where $\bm{U}$ and $\bm{V}$ are the real and imaginary parts of $\bm{Y}$, respectively. We can then define the augmented vector $\bm{\tilde{Y}}$ as:
\[
\bm{\tilde{Y}} = \begin{pmatrix} \bm{U} \\ \bm{V} \end{pmatrix},
\]
and there is a matrix $\tilde{\bm{A}} = \begin{pmatrix} \Re(\bm{A}) \\ \Im(\bm{A}) \end{pmatrix}$ such that:
\[
\bm{\tilde{Y}} = \tilde{\bm{A}}\bm{X},
\]
where $\Re(\bm{A})$ and $\Im(\bm{A})$ are the operators that extract the real and imaginary components, respectively.

The distribution of $\bm{\tilde{Y}}$ can be expressed as:
\[
\bm{\tilde{Y}} \sim \mathcal{N}\left(0, \bm{\Sigma}_{\tilde{Y}}\right),
\]
where
\scriptsize
\[
\bm{\Sigma}_{\tilde{Y}} = \begin{bmatrix} \frac{1}{2}\Re(\bm{A}\bm{\Sigma}\bm{A}^*) +\frac{1}{2}\Re(\bm{A}\bm{\Sigma}\bm{A}^T) & -\frac{1}{2}\Im(\bm{A}\bm{\Sigma}\bm{A}^*) + \frac{1}{2}\Im(\bm{A}\bm{\Sigma}\bm{A}^T) \\ \frac{1}{2}\Im(\bm{A}\bm{\Sigma}\bm{A}^*) + \frac{1}{2}\Im(\bm{A}\bm{\Sigma}\bm{A}^T) &  \frac{1}{2}\Re(\bm{A}\bm{\Sigma}\bm{A}^*) - \frac{1}{2}\Re(\bm{A}\bm{\Sigma}\bm{A}^T) \end{bmatrix}.
\]
\normalsize 

\subsection{Error plot}

\subsection{Table for Truncation number}
We investigated the truncation numbers for E (eigen-decomposition) and F (fourier transformation) and summarized the optimal values across all discretization sizes for all systems.

\label{Trun}
\begin{table}[H]
\centering
\begin{tabular}{|c|c|c|c|c|c|c|}
\hline
\textbf{Discretization} & 41 & 81 & 161 & 321 & 641 & 1281 \\ \hline
\textbf{E}              & 41 & 41 & 81  & 81 & 81  & 81   \\ \hline
\textbf{F}              & 11 & 11 & 21  & 41  & 41  & 41   \\ \hline
\end{tabular}
\caption{FitzHugh-Nagumo (FN) System}\label{tab:fn-truncation-number}
\end{table}

\begin{table}[H]
\centering
\begin{tabular}{|c|c|c|c|c|c|c|}
\hline
\textbf{Discretization} & 41 & 81 & 161 & 321 & 641 & 1281 \\ \hline
\textbf{E}              & 21 & 81 & 81  & 81  & 81  & 81   \\ \hline
\textbf{F}              & 11 & 21 & 21  & 41  & 41  & 41   \\ \hline
\end{tabular}
\caption{Hes1 System}\label{tab:hes1-truncation-number}
\end{table}

\begin{table}[H]
\centering
\begin{tabular}{|c|c|c|c|c|c|c|}
\hline
\textbf{Discretization} & 41 & 81 & 161 & 321 & 641 & 1281 \\ \hline
\textbf{E}              & 41 & 41 & 81  & 81  & 81  & 81   \\ \hline
\textbf{F}              & 21 & 21 & 41  & 41  & 41  & 41   \\ \hline
\end{tabular}
\caption{Lotka-Volterra (LV) System}\label{tab:lv-truncation-number}
\end{table}

\end{document}